\documentclass[jcp,article,accept,moreauthors,pdftex]{Definitions/mdpi}
\graphicspath{ {./figures/} }

\usepackage{listings}

\usepackage[labelformat=simple]{subfig}

\firstpage{252}
\pubvolume{1}
\issuenum{2}
\articlenumber{14}
\pubyear{2021}
\copyrightyear{2021}
\externaleditor{Academic Editor: Nour Moustafa} 
\datereceived{19 February 2021}
\dateaccepted{20 April 2021}
\datepublished{23 April 2021}
\hreflink{https://doi.org/10.3390/\linebreak  jcp1020014} 

\continuouspages{yes}

\Title{Launching Adversarial Attacks against Network Intrusion Detection Systems for IoT}
\TitleCitation{Launching Adversarial Attacks against Network Intrusion Detection Systems for IoT}

\Author{Pavlos Papadopoulos $^{1,}$*\href{https://orcid.org/0000-0001-5927-6026}{\orcidicon}, Oliver Thornewill von Essen $^{1}$, Nikolaos Pitropakis $^{1,}$*\href{https://orcid.org/0000-0002-3392-9970}{\orcidicon}, Christos Chrysoulas $^{1}$\href{https://orcid.org/0000-0001-9817-003X}{\orcidicon}, Alexios Mylonas $^{2}$\href{https://orcid.org/0000-0001-8819-5831}{\orcidicon} and William J. Buchanan $^{1}$\href{https://orcid.org/0000-0003-0809-3523}{\orcidicon}}

\AuthorNames{Pavlos Papadopoulos, Oliver Thornewill von Essen, Nikolaos Pitropakis, Christos Chrysoulas, Alexios Mylonas and William J. Buchanan}
\AuthorCitation{Papadopoulos, P.; Thornewill von Essen, O.; Pitropakis, N.; Chrysoulas, C.; Mylonas, A.; Buchanan, W.J.}
\address{%
$^{1}$ \quad Blockpass ID Lab, School of Computing Edinburgh Napier University, Edinburgh EH10 5DT, UK; 40210534@live.napier.ac.uk (O.T.v.E.);  C.Chrysoulas@napier.ac.uk (C.C.); B.Buchanan@napier.ac.uk (W.J.B.)\\
$^{2}$ \quad Department of Computer Science, University of Hertfordshire, Hatfield AL10 9AB, UK; a.mylonas@herts.ac.uk}

\corres{Correspondence: pavlos.papadopoulos@napier.ac.uk (P.P.); n.pitropakis@napier.ac.uk (N.P.)}

\abstract{\textcolor{black}{As the internet continues to be populated with new devices and emerging technologies, the attack surface grows exponentially. Technology is shifting towards a profit-driven Internet of Things market where security is an afterthought. Traditional defending approaches are no longer sufficient to detect both known and unknown attacks to high accuracy. Machine learning intrusion detection systems have proven their success in identifying unknown attacks with high precision. Nevertheless, machine learning models are also vulnerable to attacks. Adversarial examples can be used to evaluate the robustness of a designed model before it is deployed. Further, using adversarial examples is critical to creating a robust model designed for an adversarial environment. Our work evaluates both traditional machine learning and deep learning models' robustness using the Bot-IoT dataset. Our methodology included two main approaches. First, label poisoning, used to cause incorrect classification by the model. Second, the fast gradient sign method, used to evade detection measures. The experiments demonstrated that an attacker could manipulate or circumvent detection with significant probability.}}


\begin{document}

\section{Introduction}

\textcolor{black}{The internet is increasingly being populated with new devices and emerging technologies, which increases the surface of attacks, thus allowing cybercriminals to gain control of improperly secured devices. Nowadays, we are facing a more and more digitalised world, so it is more crucial than ever to detect and prevent cyber attacks as quickly as possible. Different entities ranging from homeowners to nation-states are subject to cyber-attacks, thereby increasing the challenge posed when creating mature security measures. Moreover, the~malicious parties build new skills, using more complex attack techniques to circumvent detection measures put in place~\cite{sapre2019robust}.}

\textcolor{black}{The internet is shifting towards increasing the connectivity of physical devices, called the Internet of Things (IoT), where physical objects can become active participants in business processes~\cite{sisinni2018industrial}. Increasing connectivity poses problems for businesses that need to spend more resources to protect their devices. The~security standards for IoT devices are not defined and often become an after-thought in a profit-driven market~\cite{benkhelifa2018critical}. Gartner predicts that by 2023 there will be 48.6 billion IoT devices connected using 5G~\cite{goasduff_2019}. In~addition to that, as~IoT technologies emerge, the~security risks will increase, and~traditional protection methods will become obsolete~\cite{ibitoye2019analyzing}.}

\textcolor{black}{Intrusion Detection Systems (IDS) are often put in place as a security measure to protect nodes within a network. Due to the nature of limited resources within an IoT device, IDS are often placed at the network perimeter to detect adversarial activity~\cite{soe2020towards}. One of the IoT devices' primary aim is to keep the device's processing load to a minimum. As~such, Host Intrusion Detection Systems (HIDS) are often avoided in the IoT ecosystem because of the resource-intensive activities, including file or process monitoring~\cite{elrawy2018intrusion}.}


\textcolor{black}{Machine learning models are trained using large datasets to learn patterns within the data, allowing the model to determine whether an activity is benign or not~\cite{sapre2019robust}. In~1999 the KDD-CUP-1999 dataset~\cite{cup1999data} was released and became one of the most popular datasets in research related to machine learning Network Intrusion Detection Systems (NIDS). The~KDD-CUP-99 was later updated to the NSL-KDD dataset~\cite{tavallaee2009detailed}. Over~the last two decades, both of these datasets have been losing relevance in modern IDS solutions because of the absence of modern attack methods and new protocols~\cite{nisioti2018intrusion}. However, these datasets have been defined as biased and unrealistic by multiple studies~\cite{mchugh2000testing,mahoney2003analysis,athanasiades2003intrusion}. In~2018,~ref.~\cite{koroniotis2019towards} published the Bot-IoT dataset, which was the first dataset to focus on machine learning IoT security solutions~\cite{ferrag2020deep}.}


\textcolor{black}{However, machine learning-based IDS solutions are vulnerable when the model is targeted and exploited by adversarial cybercriminals. Adversarial machine learning methods intend to cause incorrect data classification forcing the implemented model to fail~\cite{yuan2019adversarial,kurakin2016adversarial,pitropakis2019taxonomy,kantartopoulos2020exploring}. Within~an adversarial environment, it is critical to anticipate the actions an adversary may take towards an implemented model~\cite{huang2011adversarial}. To~create the best model, previous research showed the importance of using adversarial examples within a dataset when training a machine learning model~\cite{kurakin2016adversarial, xiao2015support}.}


The Bot-IoT dataset has not knowingly been created using adversarial examples towards the machine learning model. For~this reason, \textcolor{black}{our work aims} to evaluate the Bot-IoT dataset and model robustness against adversarial examples. Our contributions can be summarised as follows:

\begin{itemize}
\item We mount label noise adversarial attacks against an SVM model that detects malicious network traffic against IoT devices using the  Bot-IoT dataset~\cite{koroniotis2019towards}.
\item We generate adversarial examples using the Fast Gradient Sign Method (FGSM) against binary and multi-class Artificial Neural Networks (ANNs) using the  Bot-IoT dataset~\cite{koroniotis2019towards}.
\item \textcolor{black}{Finally, we analyse and critically evaluate the experimental results along with the model robustness against adversarial examples.}
\end{itemize}



The rest of the paper is organised into five sections, as~follows. The~literature review and background knowledge are presented in Section~\ref{chap:literature_review}. Section~\ref{chap:design_and_method} \textcolor{black}{briefly explains our methodology for the experimental activities. The~implementation and the experimental results are presented in Section~\ref{chap:implementation_results}, and~are followed with our evaluation and discussion in Section~\ref{chap:evaluation}. Finally, Section~\ref{chap:conclusions} draws the conclusion, giving some pointers for future work.}





\section{Background}
\label{chap:literature_review}
\unskip


\subsection{Intrusion Detection~Systems}
\label{sec:IDS}


\textcolor{black}{There are three popular approaches to protect networks. The~first approach is to detect an attack by setting up alert triggers, for~example, once a threshold is exceeded. Such a detection strategy informs the administrators that an attack occurs but does not prevent it. The~second approach is to prevent an attack by setting protection mechanisms that deny the attack from occurring, which raises a problem if a legitimate action is deemed illegitimate, resulting in a loss of service availability. The~final approach is to block an attack, which entails setting protections in place in retrospect of the attack, preventing or detecting the attack when it next occurs. In~the latter two approaches, the~IDS is configured as an Intrusion Prevention System (IPS) \cite{ibitoye2019analyzing}.} 


\textcolor{black}{There are two locations on which IDS can be placed depending on the sources of information; firstly, sensors can be placed on a Host system (HIDS) and secondly, sensors can be placed on the Network (NIDS) \cite{nisioti2018intrusion, van2017anomaly}. The~HIDS monitors a single host's activity such as running processes, system calls, or~code in memory~\cite{van2017anomaly, nisioti2018intrusion, benkhelifa2018critical}. The~NIDS monitors the network traffic by performing packet inspection to check protocol usage and their services or IP addresses of communication partners~\cite{van2017anomaly, nisioti2018intrusion, benkhelifa2018critical}. NIDS could commonly be placed at the gateway of ingress or egress traffic at the organisation network perimeter. Therefore, it would be possible to observe all traffic from external adversaries~\cite{oh2017security}. An~IDS is an essential tool used to identify intrusions; therefore, one of the most widespread problems in cybersecurity is creating sophisticated methods and generating effective rules for the NIDS to be beneficial~\cite{sapre2019robust}. An~IDS's success is often rated according to their accuracy values when testing controlled data of the known legitimate and illegitimate activity. The~accuracy is made up of four metrics~\cite{davis2006relationship,flach2003geometry}, and~are used when calculating the accuracy seen in Equation~(\ref{eqn:accuracy}). We have True Positive (\textit{TP}) when an attack is correctly identified and a True Negative (\textit{TN}) when a benign incident is correctly identified. When a benign incident is identified as an attack, we have a False Positive (\textit{FP}), and~when an attack is identified as a benign incident, we have a False Negative (\textit{FN}).}

\begin{equation}
\label{eqn:accuracy}
Accuracy = \frac{(TP+TN)} {(TP+TN+FP+FN)}
\end{equation}
\begin{equation}
\label{eqn:TPR}
True \; Positive \; Rate = \frac{TP} {(TP+FN)}
\end{equation}
\begin{equation}
\label{eqn:TNR}
True \; Negative \; Rate  = \frac{TN} {(TN+FP)}
\end{equation}

\begin{equation}
\label{eqn:FPR}
False \; Positive \; Rate  = \frac{FP} {(FP+TN)}
\end{equation}
\begin{equation}
\label{eqn:FNR}
False \; Negative \; Rate  = \frac{FN} {(TP+FN)}
\end{equation}


\textcolor{black}{An IDS can be classified based on its detection method into: (i) knowledge-based IDS, where the detection is based on static knowledge,  (ii) behaviour-based IDS, where the detection is based on dynamic behaviour that has been monitored, and~(iii) hybrid IDS, which is a combination of  (i) and (ii) \cite{nisioti2018intrusion}.
More specifically, a~knowledge-based IDS, also known as signature-based, is a system in which alerts are triggered based on a list of pre-determined knowledge, such as hashes, sequence of bytes, etc. A~behaviour-based IDS, also known as anomaly-based, takes on the principle of having a normal activity model, triggering alerts when anomalous activity is detected. Creating a baseline of normal activity can sometimes be inaccurate if there is malicious activity captured within the baseline. Moreover, a~representative normal activity baseline is unlikely possible with an ever-changing and expanding network~\cite{wu2019lunet, ibitoye2019analyzing}. A~behaviour-based IDS is capable of detecting zero-day attacks which have not yet been disclosed to the public~\cite{nisioti2018intrusion}. However, due to the nature of alerting all deviations from normal activity, they are known to produce high FPRs~\cite{wu2019lunet,nisioti2018intrusion}. Thirdly there is a hybrid IDS that makes use of both knowledge-based IDS and behaviour-based IDS. A~hybrid-based IDS can detect both zero-day as well as known attacks~\cite{nisioti2018intrusion}.}

\subsection{Internet of Things~Security}

\textcolor{black}{The IoT is a market that is populated with a myriad of device types. The~purpose of the IoT is to interconnect physical objects seamlessly to the information network.} IoT thrives on heterogeneity which enables the different devices to connect. As~IoT devices are often manufactured using different hardware, specific protocols and platforms must be created for these devices~\cite{oh2017security}. On~the other hand, heterogeneity limits security advancements because there is no universal international standard. Furthermore, because~of the lack of security consideration, there is some reluctance to deploy IoT devices, as~both businesses and independent parties are unwilling to risk their security~\cite{benkhelifa2018critical}.

\textcolor{black}{As IoT attacks are increasing in popularity, and~the significance of data is high (e.g., within the health sector or critical infrastructure), the~incentive to improve IoT security is increasing. However, through a profit-driven business model in a competitive market, IoT security is often considered an afterthought. Data theft is regarded as a high impact risk; however, sometimes, IoT data could be temperature monitoring, which is considered trivial unless the integrity is manipulated and fails to alert deviations.}
\textcolor{black}{Until now, the~most challenging attacks in the IoT field have been botnet attacks~\cite{soe2020towards, koroniotis2019towards}. A~botnet consists of several zombie devices which an attacker is controlling. A~large scale botnet enables a full-scale Denial of Service (DoS) attack to be carried out---such as {Mirai botnet}, or~{Bashlite}~\cite{soe2020towards}. Moreover, botnets can be used for information gathering attacks such as port-scanning and port-probing to gather information about running services on their targets~\cite{soe2020towards}.}

\textcolor{black}{Since the IoT network poses many challenges, and~there are newly introduced network varieties, traditional NIDS can no longer act independently. The~importance to protect the network lies in the pursuit of privacy. Data privacy seeks to accomplish two things. Firstly to maintain data confidentiality, and~secondly, to~uphold anonymity and to make it impossible to correlate data relations between the data and their owner~\cite{khraisat2019novel}.}

\subsection{Machine Learning Models and~Examples}
\label{sec:machine-learning}

\textcolor{black}{To assist behaviour-based IDS to reduce the number of FPRs mentioned \mbox{in Section~\ref{sec:IDS}}, researchers have demonstrated success using machine learning IDS models~\cite{soe2020towards}.} \textcolor{black}{Since a behaviour-based IDS does not rely on static knowledge, it can cope with the ever-growing network of devices and numerous zero-day attacks~\cite{atawodi2019machine, nisioti2018intrusion}.} Machine learning models are trained using a preferably large dataset, and~the intention is that the individual models recognise patterns within the data to decide whether an activity is malicious or~not.

\textcolor{black}{Machine learning models are primarily divided into three groups; firstly, supervised learning, secondly, unsupervised learning, and~finally, semi-supervised learning~\cite{atawodi2019machine, nisioti2018intrusion}. Supervised learning models are trained using labelled data where the outcome of the data is known. Labelled data might classify network traffic as benign or as an attack. On~the other hand, unsupervised learning occurs when a model is trained purely using unlabelled data. The~aim is for the model to recognise patterns within the data to classify input data. Semi-supervised learning is a mixture of supervised and unsupervised datasets, meaning that some data is labelled and other data is not. Semi-supervised learning can be useful when the dataset would be excessively time-consuming to label exhaustively.}

\textcolor{black}{Deep learning is a subset of unsupervised machine learning and intends to imitate the human brain structure through the use of neural networks. Deep learning has proven success with regards to image recognition, speech processing, and~vehicle crash prediction, among~other areas~\cite{dong2016comparison}. More recently, deep learning has shown advancements in regards to intrusion detection when compared to traditional machine learning methods~\cite{wu2019lunet}. The~success of deep learning comes through hierarchical pattern recognition within the features~\cite{van2017anomaly}.}

However, machine learning techniques are not the golden ticket towards the perfect IDS. \textcolor{black}{It is essential to use a dataset that does not have a relatively high number of redundant values because these can increase the computation time and reduce the IDS model accuracy. Moreover, the~dataset must not have unnecessary features; otherwise, the~accuracy would be reduced~\cite{khraisat2019novel}. A~significant risk of machine learning IDS being considered is using inaccurate or non-integral data to train the model. For~example, if~an adversarial insider labels malicious activity as benign, then the IDS will not detect those attack vectors, reducing the accuracy of an IDS.}

Traditional machine learning models applied to NIDS may include decision tree C4.5, Random Forests (RF), and~Support Vector Machines (SVM) \cite{dong2016comparison}. \textcolor{black}{The C4.5 machine learning model is increasingly considered out of date and is being replaced by RF models that are made up of a forest of many smaller decision trees, each specialising in different patterns.}

SVM plot each data instance within an \textit{n}-dimensional space, where \textit{n} is the number of features~\cite{koroniotis2019towards}. The~classification finds the hyperplane distinguishing classes~\cite{koroniotis2019towards}.

Deep learning models may include Artificial Neural Networks (ANN), Recurrent Neural Networks (RNN), and~Convolutional Neural Networks (CNN) \cite{ferrag2020deep}. RNNs are extended neural networks that use cyclic links, learning using memory where outputs further influence inputs for the neural network~\cite{koroniotis2019towards}. CNNs automatically generate features through rounds of learning processes~\cite{wu2019lunet}.

\textcolor{black}{A confusion matrix is the foundation towards further metrics known as precision, recall, and~F1-score. Precision is the ratio of the total number of true positives against the total number of positive predictions made, as~seen in Equation~(\ref{eqn:precision}). The~recall score is the model's ability to predict all positive values and can be seen in Equation~(\ref{eqn:recall}). The~F1-score is a weighted average of the recall and precision scores, as~seen in Equation~(\ref{eqn:f1-score}). The~final metric is the Receiving Operator Characteristic (ROC) curve and the Area Under Curve (AUC) score. The~ROC curve is plotted against the FPR and TPR, and~the AUC is the amount of space under the ROC curve. When evaluating the AUC score, it is desired to have as high of a number as possible.}

\begin{equation}
\label{eqn:precision}
Precision = \frac{TP} {(TP+FP)}
\end{equation}
\begin{equation}
\label{eqn:recall}
Recall = \frac{TP} {(TP+FN)}
\end{equation}
\begin{equation}
\label{eqn:f1-score}
F1 \; Score = 2 \times \frac{(Recall \times Precision)} {(Recall + Precision)}
\end{equation}

\subsection{Datasets}

\textcolor{black}{There are several datasets used in the IDS literature, such as the KDD-CUP-99, the~UNSW-NB15, the~CIC-DoS-2017, and~the CSE-CIC-IDS~\cite{nisioti2018intrusion}. Nonetheless, a~primary challenge of applied machine learning in IDS is the lack of publicly available and realistic datasets that can be used for model training and evaluation~\cite{koroniotis2019towards, nisioti2018intrusion}. Organisations are not willing or allowed to disclose network traffic data as they can reveal sensitive information about the network configuration and raise privacy concerns~\cite{fernandez2019deep}.}

\textcolor{black}{Moreover, several datasets have been used to study adversarial attacks in the relevant literature, such as the KDD-CUP-99, the~UNSW-NB15, the~CIC-DoS-2017 and the CSE-CIC-IDS~\cite{pacheco2021adversarial}.}
The KDD-CUP-99 dataset is the most widely researched dataset for the evaluation of IDS models~\cite{ferrag2020deep}, containing seven weeks of network traffic~\cite{cup1999data}. However, the~KDD-CUP-99 dataset is heavily criticised by researchers stating that the data is inadequate for machine learning models due to the high number of redundant rows~\cite{tavallaee2009detailed}. The~UNSW-NB15~\cite{moustafa2015unsw} collected data using various tools such as {IXIA PerfectStorm}, {Tcpdump}, {Argus}, {Bro-IDS}, in~order to identify nine attack factors Fuzzers, analysis, backdoors, DoS, exploits, generic, reconnaissance, shellcode and worms~\cite{moustafa2015unsw}. The~CIC-DoS-2017~\cite{jazi2017detecting}, is focused on DoS attacks due to the recent increase in these attacks~\cite{jazi2017detecting}. The~CSE-CIC-IDS~\cite{sharafaldin2018toward} consists of seven attack categories: {Brute-Force}, {Heartbleed}, {Botnet}, {DoS}, {Distributed Denial of Service (DDoS)}, {Web Attacks} and {Insider Attacks}.

Until recently, there have been no specific IoT datasets proposed for machine learning IDS evaluation~\cite{ferrag2020deep}. The~Bot-IoT dataset is the first to focus on an IoT network architecture and includes five different attacks; DDoS, DoS, operating system and service scans, key-logging, and~data exfiltration~\cite{koroniotis2019towards}. The~Bot-IoT dataset makes use of the Message Queuing Telemetry Transport (MQTT) protocol---a lightweight messaging protocol used in IoT architectures~\cite{ferrag2020deep}. There are over 72 million rows of data in the complete dataset; therefore, a~5\% version has also been released for more manageable use. The~dataset released with 46 features, and~the authors proposed a top-10 feature list using the Pearson Correlation Matrix and entropy values~\cite{koroniotis2019towards}. \textcolor{black}{There have been works demonstrating the efficacy of both traditional and deep learning techniques with this dataset~\cite{ibitoye2019analyzing}.}

\subsection{Related~Work}
\label{sec:advML}

\textcolor{black}{Machine learning and especially deep learning models introduce significant advantages concerning FPRs and accuracy~\cite{van2017anomaly,nisioti2018intrusion};} \textcolor{black}{however, they are not without their flaws. Machine learning models are often vulnerable to attacks with the intent to cause incorrect data classification~\cite{kurakin2016adversarial,pitropakis2019taxonomy}. Incorrectly classified data could enable an attacker to evade the IDS, thereby putting the organisation at the risk of undetected attacks. There are three primary categories of offences which include influential attacks, security violation attacks, and~specificity attacks~\cite{huang2011adversarial, xiao2012adversarial}. Influential attacks impact the model used for intrusion detection and include causative attacks that entail altering the training process by manipulating the training data and exploratory attacks that attempt to gather a deeper understanding of the machine learning model~\cite{huang2011adversarial, xiao2012adversarial}.}

\textcolor{black}{Security violation attacks target the CIA principles concerning the machine learning model~\cite{huang2011adversarial, xiao2012adversarial}. An~attacker can infer such information through analysis after gaining access to data used for model training. An~attack on model integrity aims to influence the model's metrics, such as Accuracy, F1-Score, or~AUC~\cite{yuan2019adversarial}. Consequently, if~the model's integrity cannot be maintained, the~attacker could force a model to classify data incorrectly and bypass the model's detection. An~attack on the model availability would trigger a high number of FP and FN alerts which could render a model unavailable for use. Moreover, if~there are numerous alerts, an~alert triage agent could experience alert fatigue, where the focus per alert will drop. In~turn, the~alert fatigue could lead the triage agent to dismiss a valid alert.}

Adversarial attacks in traditional machine learning are more effective at the training stage rather than the testing stage. \textcolor{black}{Attacking the model at the training stage will create fundamental changes to the machine learning model's classification accuracy.} In contrast, an~attack at the testing or deployment stage can only take advantage of inherent weaknesses in the model~\cite{biggio2011support}. An~adversary can affect either the features of a dataset (feature noise) or the labels of the dataset (label noise) \cite{biggio2011support}.
\textcolor{black}{Label noise or also referred to as label flipping, focuses on changing the labels that are used to train a model. Label flipping seems to be a more effective method of tricking a machine learning model~\cite{xiao2015support, koh2018stronger}. The~threat model of a label flipping attack is to influence the model by providing inaccurate labels during the model's training. If~the labels are not correct, then the model cannot be trained integrally and correctly~\cite{taheri2020defending}.} \textcolor{black}{There are different strategies related to label flipping, including random or targeted attacks. Random label flipping entails selecting an arbitrary percentage of labels and changing them. For~example, a~label classifying a record as malicious could be switched to benign or vice~versa. Targeted label flips focus on flipping labels that have the most significant effect on the model's performance.}
\textcolor{black}{According to~\cite{zhou2012adversarial} SVMs can often be more robust against adversarial machine learning, and~therefore a suitable choice for deployment in adversarial environments. If~attackers have access to feed incorrect labels to the machine learning model during training, then they can likely best achieve their objective~\cite{huang2011adversarial}.}

Deep learning is a subset of machine learning that shown success in classifying well-labelled data. Therefore, if~labels are not accurately assigned, then the resulting model accuracy will also be inaccurate~\cite{taheri2020defending}. \textcolor{black}{Conventional adversarial machine learning methods such as label flipping do not apply to deep learning models because the model's data is split into batches; it would be required for labels to be flipped within every batch~\cite{kurakin2016adversarial}.} 
The authors of~\cite{goodfellow2014explaining} proposed the Fast Gradient Sign Method (FGSM) to generate new perturbed data. The~FGSM is a linear manipulation of data; therefore, it is quick to generate new data~\cite{goodfellow2014explaining}. \textcolor{black}{The FGSM aims to increase the deep learning model's cost than the gradient of the loss with the input image to generate a new image.} The work in~\cite{papernot2016limitations} proposed a family of attacks under the name of Jacobian-based Saliency Map Attack (JSMA). JSMA is more computationally heavy than FGSM; however, it can create adversarial examples which appear closer to the original sample~\cite{papernot2016limitations}. The work of \cite{wiyatno2018maximal}, extended the JSMA and introduced targeted and non-targeted methods to generate adversarial examples. Nevertheless, JSMA uses more computational power than FGSM~\cite{yang2018adversarial}. 


In 2018,~ref.~\cite{wang2018deep} presented complete success in using the FGSM to get on average 26\% incorrect classification per class with their neural network model. As~reviewed \mbox{in Section~\ref{sec:advML}}, \cite{wang2018deep} used the NSL-KDD dataset and perturbed the dataset using an epsilon value of 0.02. \textcolor{black}{The experiments carried out in this paper have received a higher percentage of incorrect classification, while the epsilon value has been at least 0.1. The~result suggests that the Bot-IoT dataset and neural networks are more robust towards adversarial FGSMs than the NSL-KDD dataset.}

Nevertheless,~ref.~\cite{rigaki2017adversarial} brought up the concern that the FGSM perturbs the entire dataset of which it is given. The~impact of complete data perturbing is two-fold. Firstly, by~perturbing the Bot-IoT dataset, the~features primarily impacted are those residing within network packet headers. A~layer seven firewall could likely detect these kinds of attacks depending on the algorithm used. Secondly,~ref.~\cite{rigaki2017adversarial} propose to use the JSMA method instead of FGSM because JSMA perturbs a portion of the dataset rather than the entire dataset. On~the contrary,~ref.~\cite{yang2018adversarial} suggests that FGSM is better used in experiments because of the drawback that JSMA has with high computational power. \textcolor{black}{FGSM against the NSL-KDD was also used by~\cite{jeong2019adversarial}, who found that perturbing the reduced the accuracy of their CNN from 99.05\% to 51.35\%. Besides,~ref.~\cite{jeong2019adversarial} experimented against the NSL-KDD dataset using the JSMA method and found that the accuracy decreased from 99.05\% to 50.85\%. Comparing these results to the findings of our work's experiments, we observed a similar decrease in the accuracy, as~it can be seen in Section~\ref{chap:implementation_results}.}

\textcolor{black}{Countermeasures can be categorised into reactive and proactive~\cite{yuan2019adversarial}. Reactive countermeasures are only able to detect adversarial behaviour after the neural network has been built and deployed. 
Reactive countermeasures could be input reconstruction, and~data sanitation, which could be performed using multiple classifier systems or statistical analysis~\cite{xiao2015support, taheri2020defending}. 
Yuan~et~al. propose re-training to be a proactive countermeasure against adversaries, as~adversarial training could prepare a model in an improved manner~\cite{yuan2019adversarial}. Furthermore, if~the machine learning model is planned to be deployed in an adversarial environment, then it is crucial to use adversarial methods to train a robust model~\cite{xiao2015support, kurakin2016adversarial}. Machine learning in an adversarial environment requires analysis of attacker's methods to force inaccuracies from machine learning model~\cite{huang2011adversarial}.}

\textcolor{black}{Our work differentiates from all previous approaches because it is the first to attempt both label noise attacks and adversarial examples generation using the Bot-IoT dataset~\cite{koroniotis2019towards}. Our results support the fact that the Bot-IoT dataset combined with neural networks generates a more robust setup compared to the NSL-KDD dataset.}

\section{Methodology}
\label{chap:design_and_method}
\unskip


\subsection{Dataset~Overview}
\label{sec:dataset_description}

\textcolor{black}{There have already been many adversarial machine learning works performed on the datasets of the liteterature~\cite{pacheco2021adversarial}. However, the~Bot-IoT dataset finds itself still in its early days, and~to the best of our knowledge, no other work in the literature has used it to study adversarial machine learning. The~Bot-IoT dataset was created in a simulated environment formed up of victim and attacking machines~\cite{koroniotis2019towards}.} The traffic was captured in {.pcap} files, analysed and exported to comma-separated value ({.CSV}) files resulting more than 73~million records with 46 features, including three classification features. The work of~\cite{koroniotis2019towards}, categorised the different attacks into five classes; (1) Normal, (2) Reconnaissance, (3) DDoS, (4)~DoS, (5) Information Theft. For~the easier management of the dataset,~ref. \cite{koroniotis2019towards} extracted a random 5\% selection of each class which reduces the amount to approximately 3.6 million records. In~Table~\ref{tab:category_value_counts} the value counts of each category can be seen. The~total number of records is 73,370,443, including benign~traffic.

\vspace{-6pt}
\begin{specialtable}[H]
\small
\caption{Bot-IoT category value~counts.}
\setlength{\tabcolsep}{3.2mm}
\begin{tabular}{lllll}
\toprule
\textbf{Category} & \textbf{Full Amount} & \textbf{5\% Amount} & \textbf{Training Amount} & \textbf{Testing Amount} \\ \midrule
DDoS & 38,532,480 & 1,926,624 & 1,541,315 & 385,309 \\ \midrule
DoS & 33,005,194 & 1,650,260 & 1,320,148 & 330,112 \\ \midrule
Normal & 9543 & 477 & 370 & 107 \\ \midrule
Reconnaissance & 1,821,639 & 91,082 & 72,919 & 18,163 \\ \midrule
Theft & 1587 & 79 & 370 & 14 \\ \midrule
Total & 73,370,443 & 3,668,522 & 2,934,817 & 733,705 \\ \bottomrule
\end{tabular}
\label{tab:category_value_counts}
\end{specialtable}

In Section~\ref{sec:machine-learning}, the~importance of feature selection was emphasised. Feature selection activities were performed using the Correlation Coefficient and {Joint Entropy Score} methods against the dataset. Using the feature selection metrics, a top 10 list of features was extracted and are highlighted in Table~\ref{tab:top10_training_19_features} \cite{koroniotis2019towards}. There are three classification features in this dataset. The~first classification feature is {\textit{attack}} which is intended for binary classification. The~label is either {True} or {False} corresponding to malicious or benign traffic, respectively. All the various attack categories are labelled as True. The~second classification feature is a five-class multi-classification feature, namely {\textit{category}}, and~is made up of the string values seen in Table~\ref{tab:category_value_counts}. The~final classification feature is {\textit{subcategory}} and is composed of ten classifications. The~subcategory classification feature is more fine-grained than the category classification feature. For~example, the~DoS category is split into DoS by protocol (TCP, UDP, HTTP). Information theft was categorised into keylogging and data theft. Reconnaissance is divided into service scanning and OS fingerprinting, thus completing the ten subcategory class values. \textcolor{black}{Nevertheless, Table~\ref{tab:category_value_counts} depicts a severe data imbalance leaning towards DDoS and DoS classes than the others, something that intuitively makes sense as the IoT devices are used more frequently for such attacks as zombie devices. Besides, a~common practice in the literature to combat this asymmetry is introducing class weights during the training process~\cite{ge2019deep}.}

\begin{specialtable}[H]
\small
\caption{\textcolor{black}{Total features in the Training dataset \cite{koroniotis2019towards}.}}
\begin{tabular}{cc}
\toprule
\textbf{Features} & \textbf{Description} \\ \midrule
pkSeqID & Row Identifier \\ \midrule
Proto & \begin{tabular}[c]{@{}c@{}}Textual representation of  transaction protocols present in  network flow\end{tabular} \\ \midrule
saddr & Source IP address \\ \midrule
sport & Source port number \\  \midrule
daddr & Destination IP address \\ \midrule
dport & Destination port number \\ \midrule
attack & \begin{tabular}[c]{@{}c@{}}Class label: 0 for Normal traffic, 1 for Attack Traffic\end{tabular}  \\ \midrule
category & Traffic category  \\\midrule
subcategory & Traffic subcategory \\ \midrule
& \\ \midrule
\textbf{Top-10 Features} & \textbf{Description} \\ \midrule
seq & Argus sequence number \\ \midrule
stddev & \begin{tabular}[c]{@{}c@{}}Standard deviation of aggregated records\end{tabular} \\ \midrule
N\_IN\_Conn\_P\_SrcIP & \begin{tabular}[c]{@{}c@{}}Number of inbound connections per source IP.\end{tabular} \\ \midrule
min & \begin{tabular}[c]{@{}c@{}}Minimum duration of aggregated records\end{tabular} \\ \midrule
state\_number & \begin{tabular}[c]{@{}c@{}}Numerical representation of  transaction state\end{tabular} \\ \midrule
mean & \begin{tabular}[c]{@{}c@{}}Average duration of aggregated records\end{tabular} \\ \midrule
N\_IN\_Conn\_P\_DstIP & \begin{tabular}[c]{@{}c@{}}Number of inbound connections per destination IP.\end{tabular} \\ \midrule
drate & \begin{tabular}[c]{@{}c@{}}Destination-to-source packets   per second\end{tabular} \\ \midrule
srate & \begin{tabular}[c]{@{}c@{}}Source-to-destination packets  per second\end{tabular} \\ \midrule

max & \begin{tabular}[c]{@{}c@{}}Maximum duration of aggregated records\end{tabular} \\ \bottomrule
\end{tabular}
\label{tab:top10_training_19_features}
\end{specialtable}

\subsection{Orchestrating Attacks against Machine~Learning}
\label{sec:adv_method}

The first stage was to replicate the linear SVM model that was proposed by~\cite{koroniotis2019towards}. \textcolor{black}{The purpose of the aforementioned replicating is to enable valid comparison of metrics such as accuracy, recall, precision, F1-scores. An~artificial neural network (ANN) was implemented to train and evaluate data more quickly than the RNN and LSTM models.} Nevertheless, the~activation functions and neural network structure remained the same for a fairer comparison. The~second and third stages focused on adversarial examples for both the SVM and ANN. With~SVM adversarial examples, the~undertaken method was labelled noise generation during the training phase. \textcolor{black}{As previously discussed, the label noise generation is a more effective attack against traditional machine learning models than feature noise experiments.} Firstly, a~certain percentage of data labels was manipulated, ranging from 0\% to 50\% of flipped labels. The~SVM model was trained using adversarial examples. The~increasing rate of flipped labels should incrementally reduce the accuracy of the SVM models. \textcolor{black}{However, as~discussed in Section~\ref{sec:advML}, random label manipulation could also result in little effect on model accuracy. Accordingly, targeted label manipulation was also performed. On~the default SVM model, each label's margin to the SVM hyperplane is calculated; consequently, the~top selected percentage (ranging from 0\% to 50\%) with the smallest margin to the SVM hyperplane was chosen, and~their label is flipped. Using the targeted label flipping method should result in a more significant impact on model accuracy and negatively impact the metrics.}

After experimenting with SVM manipulation, FGSM methods were applied to the data used for the ANN. \textcolor{black}{In contrast to the SVM threat model, which assumes that the attacker can manipulate the training data, the~FGSM assumes that the attacker controls the data and tries to evade the model after it has been deployed. The~{CleverHans} package was used to generate adversarial examples based on the testing dataset, which simulates attacking the model after the deployment~\cite{papernot2018cleverhans}. As~discussed in Section~\ref{sec:advML}, the~noise factor determines the level at which the generated adversarial examples will differ from the non manipulated data.} The work of \cite{wang2018deep}, found that a noise factor of 0.02 was sufficient to get 100\% incorrect classification on the NSL-KDD dataset. However, as~the NSL-KDD dataset is smaller in size than the Bot-IoT dataset, a~noise factor from 0 to 1, with~an increment in steps of 0.1, was applied and evaluated. The~ANN was trained on a binary classification model, similar to that of the RNN and LSTM models proposed by~\cite{koroniotis2019towards}. However, neural networks have proven to be highly successful using multi-class classification. As~the Bot-IoT dataset also contains five multi-class label features, experimentation using FGSM was carried out in multi-classification~mode.

\subsection{Evaluation~Criteria}
\label{sec:eval_criteria}

\textcolor{black}{The fourth and final stage is to evaluate the results, comparing adversarial examples to the results without any data manipulation. The~accuracy, precision, recall, and~F1-scores were compared to evaluate both the SVM and ANN models. In~addition to these numeric metrics, confusion matrices were generated, which graphically represent the incorrect classifications. Further, as~explained in Section~\ref{sec:machine-learning}, the~recall score is a key-metric as a decrease shows the extent that the model classifies false negatives. An~increased count of FNs could allow a cyber attacker to launch attacks which the machine learning or deep learning models would not detect.}

\section{Implementation and~Results}
\label{chap:implementation_results}
\unskip


\subsection{Data~Preparation}
\label{sec:dataset_preparation}

The first stage of data preprocessing is to reduce the number of features to increase accuracy and decrease the chance of overfitting models. Ref.~\cite{koroniotis2019towards} used the Correlation Coefficient and the Joint Entropy Scores to extract the top-10 features of this dataset, as~described in Table~\ref{tab:top10_training_19_features}. For~more manageable model development,~ref.~\cite{koroniotis2019towards} have excerpted 5\% of the total records resulting in 3.6 million records. \textcolor{black}{\cite{koroniotis2019towards} have also already split the 5\% extracted records into training and testing splits, sized 80\% and 20\% respectively.}
The released datasets include additional features which could be useful for further statistical analyses; making a total of 19 features. The~full list of features, their description and the top-10 features can be seen in Table~\ref{tab:top10_training_19_features}. The~features {proto}, {saddr}, {sport}, {daddr}, {dport} can be used to identify data points uniquely and are therefore considered flow identifiers and are removed for model development~\cite{koroniotis2019towards}. The~{pkSeqID} feature is a row identifier and is consequently also removed for model~development.

The training features did not need to be encoded as the data were already in numeric form. However, the~{category} feature used for five-class multi-classification must be encoded using Scikit-Learn's preprocessing OneHotEncoder function, which transforms categorical features into a numerical array. As~the margin of labels to the hyperplane will be measured and used for label manipulation, the~data must be normalised. The~purpose of data normalisation is to apply an equal scale to the training and testing splits without distorting the ranges within the relative values. Scikit-learn's preprocessing MinMaxScaler can be used to scale the data between values of $-$1 and~1.



After pre-processing the data, it was essential to train a machine learning model using trusted datasets. \textcolor{black}{The trusted dataset means that there has been no data manipulation performed yet, which ensures data integrity. Using data that maintains integrity allows the creation of models which perform at their best. Besides, the~trusted models can be compared to the models that undergo adversarial training or testing manipulation.}

\subsubsection{SVM Trusted~Model}

\begin{figure}[H]
\begin{tabular}{cc}
\includegraphics[width=0.4\linewidth]{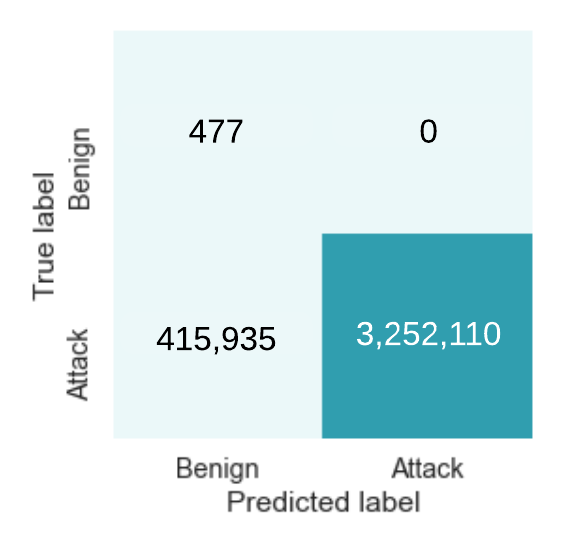}
&\includegraphics[width=0.5\linewidth]{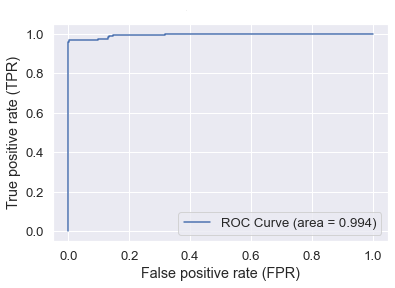}\\
({\bf a})&({\bf b})\\
\end{tabular}


\caption{(\textbf{a}) {Support Vector Machines confusion matrix without label flipping}. (\textbf{b}) {Support Vector Machines Receiving Operator Characteristic curve without label flipping}.}
\label{fig:default_svm}
\end{figure}

The trained SVM model is a Support Vector Classifier (SVC) using the linear kernel. The~hyper-parameters were primarily the default parameters in the Scikit-Learn package, except~for the penalty score set to 1 and the maximum iterations set to 100,000 as replicated by~\cite{koroniotis2019towards}. Rather than evaluating the model using the training and testing split and as explained in Section~\ref{sec:machine-learning}, four-fold cross-validation was used. \textcolor{black}{Similar to the method of~\cite{koroniotis2019towards}, the~combined training and testing datasets were used to generate the confusion matrix. The~generated confusion matrix can be seen in Figure~\ref{fig:default_svm}a, and~the ROC curve in Figure~\ref{fig:default_svm}b. Viewing the confusion matrix, the~model inaccurately predicts 415,935 true attack traffic as predicted benign traffic.} The ROC curve and, consequently, the~AUC score show that the model was highly capable of detecting attack traffic over benign traffic. Further, the~accuracy, recall, precision and F1 scores can be seen in Table~\ref{tab:temps}.

\vspace{-6pt}
\unskip

\subsubsection{ANN Trusted~Model}

\begin{figure}[H]
\includegraphics[width=0.6\linewidth]{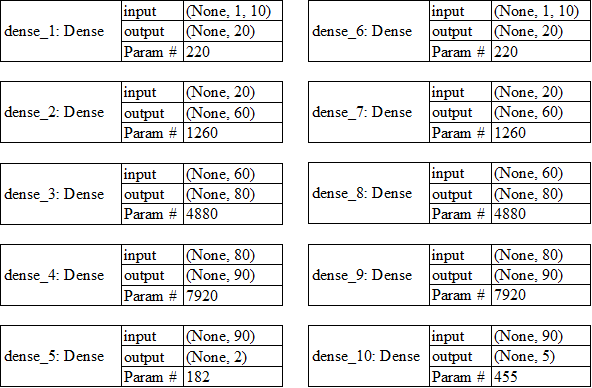}
\caption{Artificial Neural Networks~Design.}
\label{fig:ann_design}
\end{figure}

The trained ANN had one input layer, three intermittent layers and one output layer. The~input layer was composed of ten nodes, equally to the number of features. The~intermittent layers are 20, 60, 80, 90, respectively. The~output layer was either two for binary classification or five for multi-class classification. The~activation function of the intermittent layers are {TanH}, and~the output activation layer was {Sigmoid}. \textcolor{black}{Sigmoid was chosen for the output layer to replicate the~\cite{koroniotis2019towards} method because Sigmoid models are more robust~\cite{goodfellow2014explaining}.} The ANN designs for both binary and five-class classification layers can be seen \mbox{in Figure~\ref{fig:ann_design}}. The~left side shows the ANN design for the binary classification, while the right side shows the ANN design for multi-class~classification.

\begin{figure}[H]

\begin{tabular}{cc}
\includegraphics[width=0.4\linewidth]{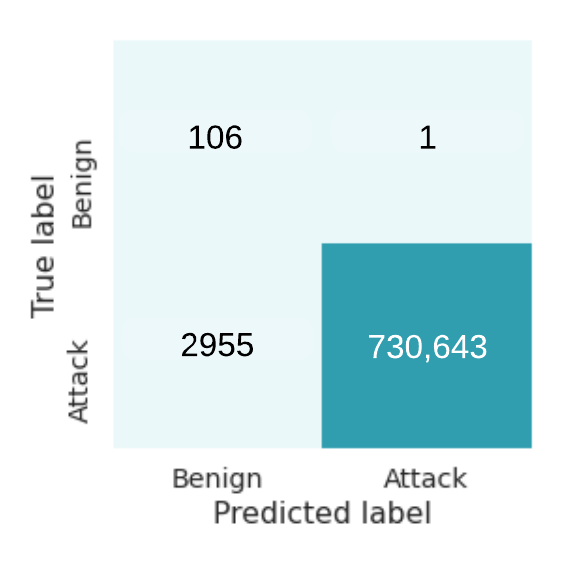}
&\includegraphics[width=0.4\linewidth]{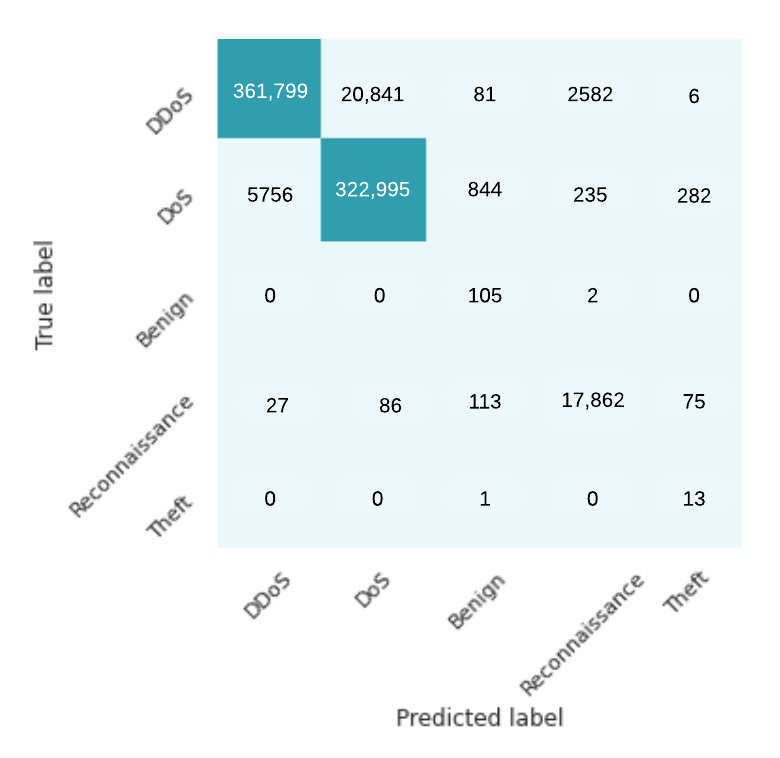}\\
({\bf a})&({\bf b})\\
\end{tabular}

\caption{(\textbf{a}) {Binary Artificial Neural Networks confusion matrix without adversarial examples}. (\textbf{b})~{Multi-class Artificial Neural Networks confusion matrix without adversarial examples}.}
\label{fig:default_ann_conf_mat}
\end{figure}

\textcolor{black}{The confusion matrices for the ANN were calculated using the testing dataset. The~testing dataset is used rather than the full dataset with the SVM model because there is a different threat model. As~stated in Section~\ref{sec:adv_method}, the~ANN's threat model was that the attacker can manipulate data after the model has been deployed. The~testing dataset was used to simulate adversarial examples by an attacker. The~binary classification confusion matrix in Figure~\ref{fig:default_ann_conf_mat}a, shows that the model  incorrectly classified 2955 true attack records as benign. In~Figure~\ref{fig:default_ann_conf_mat}b, the~confusion matrix for five-class multi-classification shows that the model inaccuracies lie in predicting more labels as DDoS or DoS compared to their true labels. Further, Table~\ref{tab:temps} shows that the ANN can report very high accuracy, recall, precision, and~F1 scores while maintaining a low loss score.}

\begin{specialtable}[H]
\caption{SVM and ANN scores without label flipping and adversarial~examples.}
\scalebox{.745}[0.745]{\begin{tabular*}{\textwidth}{@{\extracolsep{\fill}}ll}
\toprule
{(a) {SVM scores without label~flipping}}  &\\ \midrule
\textbf{Scoring} & \textbf{Percentage (\%)} \\ \midrule
Accuracy & 85.897 \\ \midrule
Recall & 85.895 \\ \midrule
Precision & 100 \\ \midrule
F1 & 91.255 \\  \midrule
{(b) ANN scores without adversarial~examples} &\\ \midrule
\textbf{Scoring} & \textbf{Percentage (\%)} \\ \midrule
Accuracy & 99.692 \\ \midrule
Loss & 1.170 \\ \midrule
Recall & 99.813 \\ \midrule
Precision & 99.591 \\ \midrule
F1 & 99.702 \\ \bottomrule
\end{tabular*}}
\label{tab:temps}
\end{specialtable}

\subsection{Creating Adversarial~Examples}
\label{sec:creating_adv_examples}

\textcolor{black}{This subsection details the undertaken activities to generate adversarial labels and is split into two parts.} The SVM part details how to perform random and targeted label flipping on the training dataset. The~ANN part specifies how to use the CleverHans package to generate adversarial examples~\cite{papernot2018cleverhans}. The~CleverHans FGSM also implements a targeted and non-targeted approach. The~targeted FGSM approach aims to alter the dataset to appear similar to the classification feature. On~the contrary, the~non-targeted FGSM approach aims to modify the labels of the~dataset.

\subsubsection{Support Vector Machine Label Flipping~Activities}
\label{sec:SVM_flip_activities}

\textcolor{black}{As stated in Section~\ref{sec:adv_method}, there are two approaches to flip the labels: random label flip and targeted label flip. Additionally, the~number of labels to be flipped ranges from 5\% to 50\%, in~5\% increments (5\%, 10\%, {\ldots}, 45\%, 50\%) (Appendix \ref{app:SVMrandomflip} displays the method to flip a random percentage of labels). The~first stage is to sample the training dataset, in~this case, 5\% (as defined by n), and~store the indexes of this sample.} The second stage is to alter the {\textit{attack}} column and use a binary {Exclusive OR} operation to flip the label from 1 to 0 and vice~versa. After~flipping the labels, data must be pre-processed with the same steps seen in Section~\ref{sec:dataset_preparation}. Placing this code within a loop allows one to iteratively increase the percentage of labels manipulated and further and train the model with the modified~dataset.

\textcolor{black}{Prior to executing the SVM targeted label flip, each label's distance from the hyperplane requires to be measured. The~distance to the hyperplane is known as the margin. As~explained in Section~\ref{sec:adv_method}, the~records with the lowest margin should have the highest impact on the metrics.} The impact is high because these labels affect the orientation of the hyperplane when a model is trained (Appendix \ref{app:SVMtargetedflip}, shows a function that can be used to select the nearest \textit{X\%} record indexes). Upon~retrieving the relevant indexes, the~defined function applies a binary {Exclusive OR} operation to flip the label from 1 to 0 and vice~versa.

\subsubsection{Creating Artificial Neural Network Adversarial~Examples}

The activities to create adversarial models using the FGSM are the same for the binary and the five-class multi-classification ANNs. \textcolor{black}{The difference between the binary and five-class multi-classification lies in the feature column; hence, after~the pre-processing stage, the~data formatted correctly and is ready for the CleverHans 3.0.1 package~\cite{papernot2018cleverhans}.}

The first stage of using the CleverHans package includes converting the Keras model into a logistic regression model (Appendix \ref{app:logits}). Following this, the~testing dataset must be converted into a Tensorflow Tensor (Appendix \ref{app:converttensor}). After~converting the testing dataset into a Tensor, an~epsilon value must be defined as it determines the scale of which the data is perturbed. Using the \texttt{fast\_gradient\_method} import, it is possible to generate a new testing dataset (Appendix \ref{app:fgsm}, line 6 defines whether the FGSM is a targeted attack or not). The~generated new testing dataset is used to evaluate the ANN model and a confusion matrix as well as the accuracy, loss, recall, precision, and~F1 metrics are~recorded.

\subsection{Adversarial Example~Results}


\subsubsection{Support Vector Machine Label Flipping~Result}

As described in Section~\ref{sec:SVM_flip_activities}, both a random and targeted selection of labels ranging from 0\% to 50\% were flipped. It was hypothesized that the targeted label manipulation should present a greater effect on how the SVM calculates its hyperplane. \textcolor{black}{Appendix~\ref{app:metricsplots}, Figure~\ref{fig:svm_targeted_non_targeted}a shows that as the number of labels flipped increased to 50\%, all of the scores decreased, showing success in label flipping towards manipulating the SVM. Nonetheless, looking at the metric gradients, it can be observed that the accuracy increased in some cases. The~increase in accuracy is surprising as the increased number of label manipulation should cause the accuracy to decrease continuously. An~example of accuracy increasing is seen from 20\% to 25\% of labels flipped.}

Performing the random label manipulation activities, it was expected that the metrics could be influenced significantly or minimally depending on the random sample of labels selected.\textcolor{black}{ We assumed that the random sample selection included labels that had a small margin to the default hyperplane. In~that case, the~impact should be more significant compared to if the labels had a large hyperplane margin.} \textcolor{black}{The metrics from the experiments carried out are presented in Appendix~\ref{app:metricsplots}, Figure~\ref{fig:svm_targeted_non_targeted}b.} \textcolor{black}{The metrics confirm the hypothesis and show eradicate metrics; there is no correlation to the increasing label manipulation percentage and the metrics retrieved.} Table~\ref{tab:svm_label_flip_min_max} contains the numerical metrics with 0\% and 50\% label~manipulation.


\begin{specialtable}[H]
\small
\caption{Effect of Zero vs. 50\% label flips against the metrics using hyperplane margin~method.}
\setlength{\tabcolsep}{2.2mm}
\begin{tabular}{lcccccccc}
\toprule
\textbf{Scoring} & \multicolumn{2}{c}{\textbf{Accuracy}} & \multicolumn{2}{c}{\textbf{Precision}} & \multicolumn{2}{c}{\textbf{Recall}} & \multicolumn{2}{c}{\textbf{F1}} \\ \midrule
Percentage of flipped labels (\%) & 0 & 50 & 0 & 50 & 0 & 50 & 0 & 50 \\ \midrule
Random Flip & 0.999 & 0.441 & 0.999 & 0.610 & 1.0 & 0.613 & 0.999 & 0.612 \\ \midrule
Targeted Flip & 0.999 & 0.610 & 0.999 & 0.621 & 1.0 & 0.913 & 0.999 & 0.737 \\ \bottomrule
\end{tabular}
\label{tab:svm_label_flip_min_max}
\end{specialtable}

\subsubsection{Artificial Neural Network Adversarial Examples~Result}

\textcolor{black}{In Appendix~\ref{app:metricsplots}, Figure~\ref{fig:binary_non_targeted_targeted}b, the~targeted FGSM is applied to the binary classification problem. The~accuracy, precision, F1, and~recall scores were reduced as the epsilon value grew.} Additionally, the~loss score increased as the epsilon value increases. However, the~accuracy and precision scores are insignificantly impacted until the epsilon value is 0.8. The~recall and precision scores are significantly impacted, with~a lower epsilon score of 0.5. Table~\ref{tab:fgsm-epsilon-min-max} contains the numerical metrics of zero epsilon as well as 1 epsilon values. \textcolor{black}{Viewing Figure~\ref{fig:binary_non_targeted_targeted}a which is the non-targeted FGSM to the binary classification problem, it is seen that the attack is more significant compared to the targeted attack. Figure~\ref{fig:binary_non_targeted_targeted}a shows that the accuracy, precision, recall, and~F1 scores are all impacted when the epsilon value is 0.2 and higher.}

\begin{figure}[H]
\widefigure
\begin{tabular}{cc}
\includegraphics[width=0.45\linewidth]{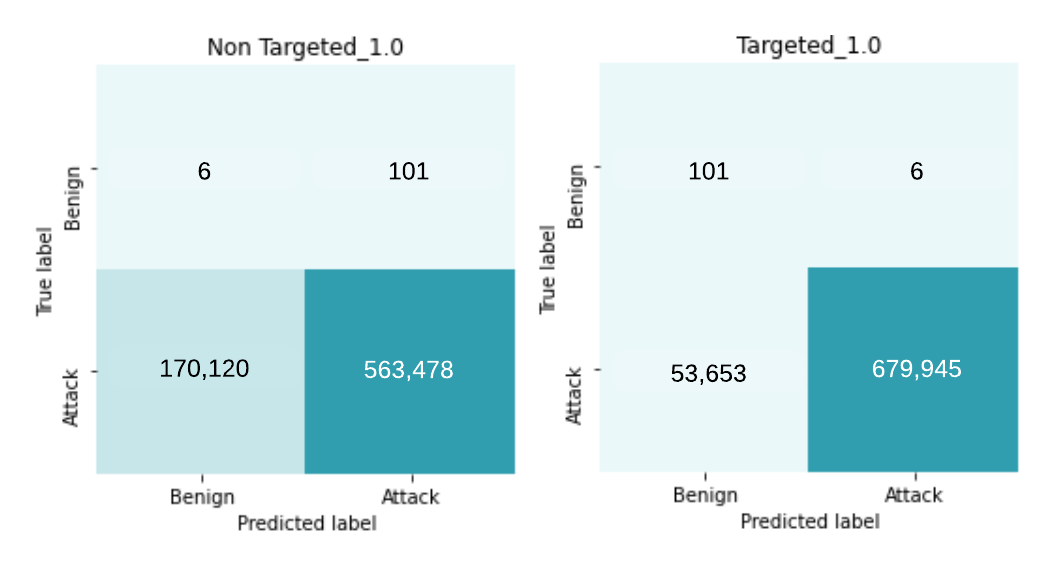}
&\includegraphics[width=0.45\linewidth]{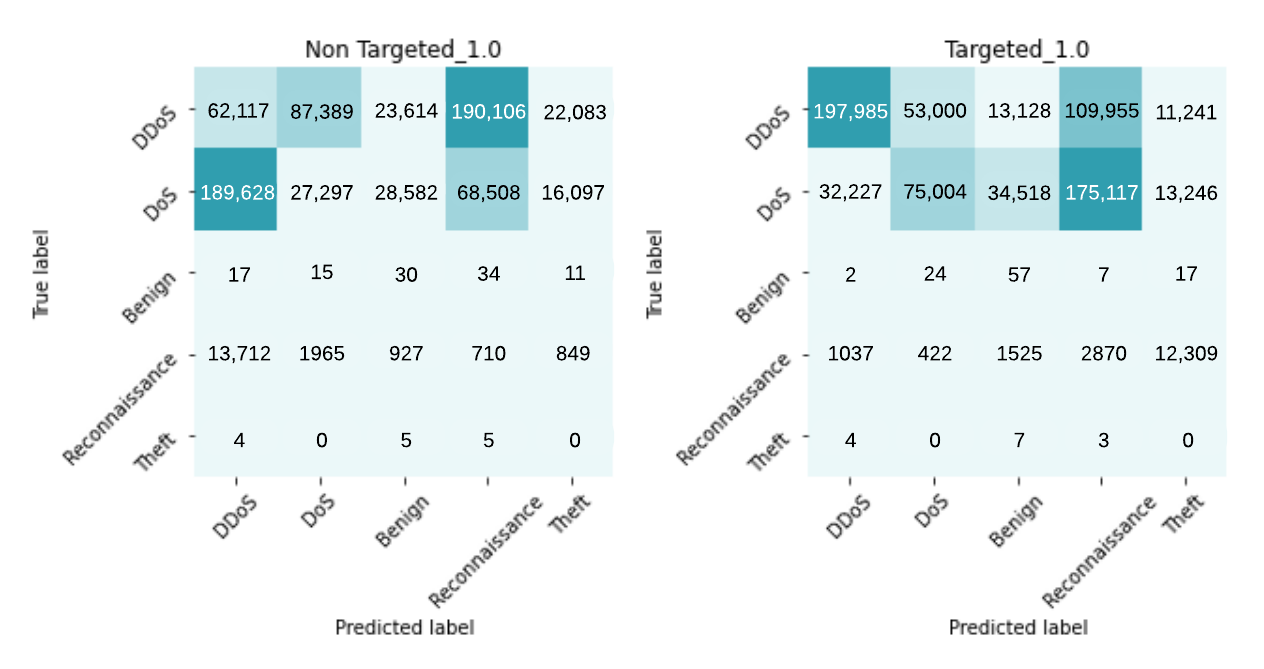}\\
({\bf a})&({\bf b})\\
\end{tabular}

\caption{(\textbf{a}) {Confusion matrices of binary class perturbed with 1.0 epsilon}. (\textbf{b}) {Confusion matrices of multi-class perturbed with 1.0 epsilon}.}
\label{fig:fgsm_binary_1_multi_1}
\end{figure}

The Confusion matrices in Figure~\ref{fig:fgsm_binary_1_multi_1}a are an alternative representation of the effect that a 1.0 epsilon value had. On~the left in Figure~\ref{fig:fgsm_binary_1_multi_1}a is the non-targeted method, and~on the right is the targeted method. In~the non-targeted method, it is seen that there is a greater amount of incorrectly predicted benign records. \textcolor{black}{The greater amount of false-negative cases affects lowering the recall score significantly, as~confirmed in Appendix~\ref{app:metricsplots}, Figure~\ref{fig:binary_non_targeted_targeted}b.} \textcolor{black}{On the other hand, viewing Figure~\ref{fig:multi_targeted_non_targeted}b, the~targeted FGSM attack on the multi-class ANN model shows that the accuracy, precision, F1, and~recall scores gradually decrease while the epsilon value grows to 1. Interestingly, in~the non-targeted FGSM applied to the multi-class ANN model, the~accuracy, precision, F1 and recall scores sharply fall until an epsilon value of 0.1, though~the impact thereon is insignificant.}

The confusion matrices shown in Figure~\ref{fig:fgsm_binary_1_multi_1}b were created using the perturbed testing dataset, with~an epsilon value of 1.0 in the multi-class ANN model. \textcolor{black}{On the left side, the~non-targeted FGSM is presented, while the targeted FGSM is presented on the right. Comparing these confusion matrices with the baseline multi-class confusion matrix in Figure~\ref{fig:default_ann_conf_mat}b, it is observed that the model is lowering its capability of detecting DDoS and DoS attacks while incorrectly predicting a higher number of benign, reconnaissance, and theft~attacks.}


\end{paracol}
\nointerlineskip
\newpage

\vspace{-9pt}
\begin{specialtable}[H]
\widetable

\caption{Effect of Zero epsilon vs. 1 epsilon against the metrics using Fast Gradient Sign Method in targeted and non-targeted~modes.}
\setlength{\tabcolsep}{3.3mm}
\begin{tabular}{lcccccccccc}
\toprule
\textbf{Scoring} & \multicolumn{2}{c}{\textbf{Accuracy}} & \multicolumn{2}{c}{\textbf{Loss}} & \multicolumn{2}{c}{\textbf{Precision}} & \multicolumn{2}{c}{\textbf{Recall}} & \multicolumn{2}{c}{\textbf{F1}} \\ \midrule
Epsilon & 0 & 1 & 0 & 1 & 0 & 1 & 0 & 1 & 0 & 1 \\ \midrule
Binary Targeted & 0.996 & 0.927 & 0.016 & 0.151 & 0.996 & 0.895 & 0.996 & 0.563 & 0.996 & 0.690 \\ \midrule
Binary Non-Targeted & 0.996 & 0.768 & 0.016 & 1.080 & 0.996 & 0.769 & 0.996 & 0.771 & 0.996 & 0.769 \\ \midrule
Multi-Targeted & 0.956 & 0.421 & 0.045 & 1.764 & 0.952 & 0.312 & 0.957 & 0.493 & 0.955 & 0.382 \\ \midrule
Multi-Not-Targeted & 0.956 & 0.141 & 0.045 & 2.403 & 0.952 & 0.153 & 0.957 & 0.249 & 0.955 & 0.189 \\ \bottomrule
\end{tabular}
\label{tab:fgsm-epsilon-min-max}
\end{specialtable}
\vspace{-9pt}
\begin{paracol}{2}
\switchcolumn

\section{Evaluation and~Discussion}
\label{chap:evaluation}
\unskip


\subsection{Machine Learning Model~Metrics}

When training different models, one must consider the metrics that can be used when comparing them. Primarily, the~focus is to achieve a high level of accuracy while also retaining low FPRs~\cite{fernandez2019deep}. \textcolor{black}{A confusion matrix can also be used with labelled datasets to visualise the values used to calculate the accuracy.}
\textcolor{black}{It is possible to exhaustively evaluate the various machine learning models through a combination of the varying metrics.} The evaluation stage of an IDS model is highly critical as it can help direct further improvements for the model during the testing phase. Moreover, a~holistic evaluation can help to select the most applicable IDS model for~deployment.

\subsection{SVM Dataset Label~Manipulation}

In Section~\ref{sec:eval_criteria}, it was discussed that an increase in FNR would allow for a greater likelihood of the model not detecting malicious events. \textcolor{black}{The FNR is inversely proportional to the recall score; albeit the FN count increases, the~recall score will decrease.} \textcolor{black}{The recall metrics that are generated in Appendix~\ref{app:metricsplots}, Figure~\ref{fig:svm_targeted_non_targeted} have been extracted, and~are seen in Figure~\ref{fig:svm_recall_scores}.} \textcolor{black}{It is observed in Figure~\ref{fig:svm_targeted_non_targeted}a that in targeted label flipping, the~accuracy score falls at a quicker rate than the recall score in the case of targeted label flipping. This is because the number of FN is not increasing as quickly as the false positive rate. In~the case of random label flipping, as~observed in Figure~\ref{fig:svm_targeted_non_targeted}b, the~recall score is firmly related to the accuracy score. The~strong correlation could mean that the accuracy is falling because the FNR increases more than the targeted experiment. Therefore, the~outcome is that an adversary would have more success if they performed random label flipping. However, comparing these results to the works of~\cite{xiao2015support,koh2018stronger}, it was found that one must flip labels with a large margin to distort the hyperplane significantly. The~hypothesis suggested in Section~\ref{sec:adv_method} of flipping labels with a small margin to the hyperplane is false. To~distort the hyperplane, labels' generation with large margins must be selected.}

\vspace{-6pt}
\begin{figure}[H]
\includegraphics[width=0.6\linewidth]{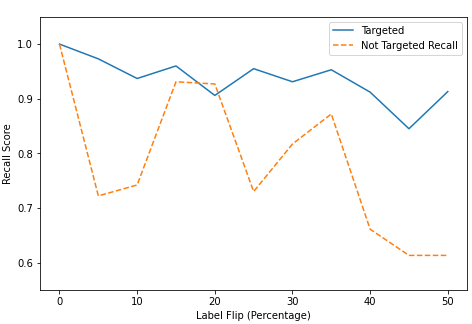}
\caption{Comparing recall scores in targeted/non-targeted Support Vector Machines label~flip.}
\label{fig:svm_recall_scores}
\end{figure}

\textcolor{black}{Still, further concerns must be considered regarding the targeted label flipping attack against the SVM.} The metrics show a decreasing trend as the percentage of label flips go up. However, it must be considered why the metric gradient is repetitively alternating from positive to negative. \textcolor{black}{The complete effect of 0\% and 50\% label flip in Table~\ref{tab:svm_label_flip_min_max} shows that the metrics all decrease, nevertheless in Appendix~\ref{app:metricsplots}, Figure~\ref{fig:svm_targeted_non_targeted} shows the alternating positive and negative metric gradients.} This anomaly could be explained depending on how the hyperplane is adjusted after flipping \textit{X\%} of labels. For~example, labels with a margin of 1\% over those flipped labels may be manipulating the result. Therefore, the~alternating positive and negative metric gradients is an~anomaly.

Finally, the~method of attack against the SVM must be considered. The~threat model used against the SVM assumed that the adversary had access to the training dataset. \textcolor{black}{The authors in~\cite{huang2011adversarial} stated that if an adversary can manipulate the training dataset, the~results should correspond with a strong negative impact on the metrics. Nevertheless, even though the accuracy and recall scores decreased, the~metrics' amount could be considered insignificant. It took a large amount of data manipulation to substantially impact the metrics, using adversarial countermeasures such as statistical analysis, and~this attack could be detected~\cite{xiao2015support}.}

\subsection{ANN Adversarial Example~Generation}

In Section~\ref{sec:dataset_description}, it was described that the dataset could be used to train neural networks with a multi-classification problem. The~experiments were carried out, generating adversarial examples using the FGSM against binary and multi-classification ANN models. The~threat model assumed that the adversary could manipulate the data after the model had been deployed. An~adversarial attack was simulated by manipulating the testing dataset. \textcolor{black}{The CleverHans package can generate targeted, and~non-targeted adversarial examples using the FGSM~\cite{papernot2018cleverhans}.} Both targeted and non-targeted FGSMs tested with the binary and multi-class ANN models. In~Appendix~\ref{app:metricsplots}, the~Figures~\ref{fig:binary_non_targeted_targeted} and \ref{fig:multi_targeted_non_targeted} show that generating adversarial examples negatively impacted the metrics. The~negatively impacted metrics show that the intrusion detection system failed to detect adversarial examples. Therefore an adversary could use the FGSM to bypass the machine learning IDS~model.

\subsubsection{Binary Classification ANN~Model}

Using the targeted FGSM to generate adversarial examples showed decreased accuracy when the epsilon value reached 0.8. As~explained in Section~\ref{sec:eval_criteria}, a~decreased recall metric shows an increase in false negatives, enabling an attacker to bypass the IDS. Figure~\ref{fig:fgsm_recall_vs_accuracy}a shows that as the epsilon value increases to 1, the~accuracy falls from 99.8\% to 92.7\%, and~the recall score falls from 99.7\% to 56.3\%. In~addition, the~confusion matrix seen in Figure~\ref{fig:fgsm_binary_1_multi_1}a, shows a significant increase in false-negative classifications than the original confusion matrix in Figure~\ref{fig:default_ann_conf_mat}a. The~learning outcome is that while the accuracy remains high, the~recall score falls significantly. For~example, if~an organisation accepts a model with over 90\% accuracy, then they are at a greater risk of the success of an undetected attack using the FGSM. Using the non-targeted FGSM to generate adversarial examples showed a more significant negative impact than the targeted FGSM. In~Figure~\ref{fig:fgsm_recall_vs_accuracy}b, the~accuracy score fell from 99.6\% to 76.8\%, and~the recall score fell from 99.6\% to 77.1\%. \textcolor{black}{It was noted in Section~\ref{sec:creating_adv_examples} that the CleverHans package could perform a targeted and non-targeted attack. The~non-targeted attack focuses on trying to make the labels incorrect~\cite{papernot2018cleverhans}. The~attack on the labels forces the model to incorrectly classify any label, which could be why the accuracy is falling more significantly compared to the targeted attack.}

\vspace{-6pt}
\begin{figure}[H]

\begin{tabular}{cc}
\includegraphics[width=0.4\linewidth]{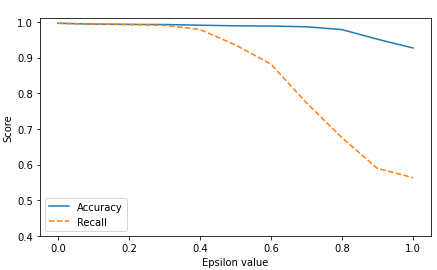}
& \includegraphics[width=0.4\linewidth]{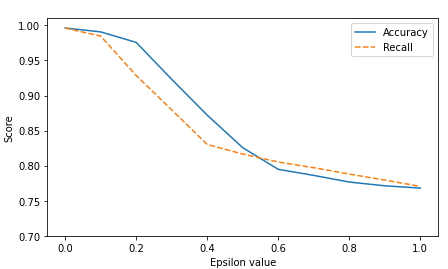}\\
({\bf a})&({\bf b})\\
\end{tabular}

\caption{(\textbf{a}) {Comparing accuracy versus recall scores in binary targeted Fast Gradient Sign Method}. (\textbf{b}) {Comparing accuracy versus recall scores in binary non-targeted Fast Gradient Sign Method}.}
\label{fig:fgsm_recall_vs_accuracy}
\end{figure}

However, taking the confusion matrices in Figure~\ref{fig:fgsm_binary_1_multi_1}a into consideration, it is observed that the non-targeted FGSM created adversarial examples results in a higher number of false negatives. \textcolor{black}{The recall score falls significantly in the targeted FGSM and seen in \mbox{Figure~\ref{fig:fgsm_recall_vs_accuracy_multi}a} because the model's ability to classify true positives correctly remains high.}

\subsubsection{Multi-Class ANN~Model}

\textls[-10]{\textcolor{black}{Using the targeted FGSM against the multi-class ANN showed a gradual negative impact during the time the epsilon value grew to 1, as~seen in Figure~\ref{fig:fgsm_recall_vs_accuracy_multi}a. The~accuracy score fell from 95.6\% to 42.1\%, and~the recall score fell from 95.7\% to 49.3\%.} The targeted FGSM aims to move the dataset in the direction of the feature class. In~the non-targeted FGSM activities, the~metrics had a greater impact. The~metrics in the non-targeted FGSM fall sharply with an epsilon of 0.1, but~insignificantly, thereafter, as~seen in Figure~\ref{fig:fgsm_recall_vs_accuracy_multi}b. \textcolor{black}{The recall score falls significantly in the non-targeted FGSM; however, it must be considered that an adversary might not bypass the machine learning ANN IDS. Hence, as~the false-negative rate increases, there is still a significant chance that an attack is detected but incorrectly labelled as another attack, thereby creating an alert. Nevertheless, the~increased false-negative rate will create more overhead for those within an organisation that triage the~alerts.} }
\vspace{-4pt}

\begin{figure}[H]

\begin{tabular}{cc}
\includegraphics[width=0.4\linewidth]{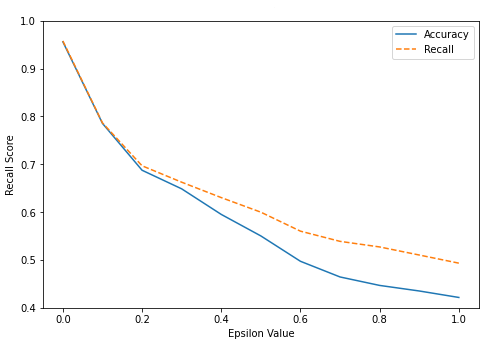}
& \includegraphics[width=0.39\linewidth]{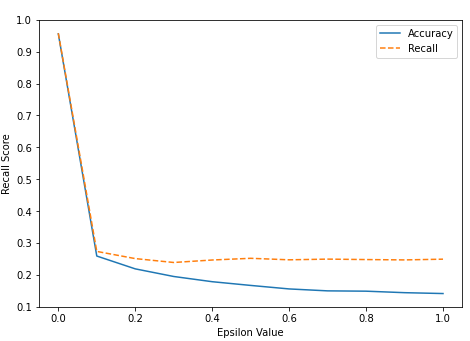}\\
({\bf a})&({\bf b})\\
\end{tabular}

\caption{(\textbf{a}) {Comparing accuracy versus recall scores in multi-class targeted Fast Gradient Sign Method}. (\textbf{b}) {Comparing accuracy vs recall scores in multi-class non-targeted Fast Gradient Sign Method}.}
\label{fig:fgsm_recall_vs_accuracy_multi}
\end{figure}

\textls[-0]{\textcolor{black}{The confusion matrices of the ANN after applying the FGSM to the testing dataset in Figure~\ref{fig:fgsm_binary_1_multi_1}b, to~the confusion matrix without data manipulation in Figure~\ref{fig:default_ann_conf_mat}b, it is seen that the model is predicting a large portion of the DDoS and DoS traffic as benign.} A small part of reconnaissance attacks were classified incorrectly as benign, and~interestingly theft traffic was classified with a similar accuracy degree. \textcolor{black}{The takeaway is that an adversary could launch DDoS and DoS more successfully, whilst data theft attacks would still be~detected.} }

\section{Conclusions}
\label{chap:conclusions}

\textcolor{black}{The study of IDS has become a rapidly growing field once more due to the involvement of machine learning models.} \textcolor{black}{Machine learning-based IDS are vulnerable to adversarial attacks, in~which the intent is to cause incorrect data classification so adversaries can avoid detection.}
\textcolor{black}{Our work aims to provide a deeper understanding of adversarial machine learning methods and how they can bypass machine learning-based IDS. Our work evaluated the robustness of the Bot-IoT dataset
against label noise attacks using an SVM model. In~addition to the SVM models, adversarial example generation was experimented using the FGSM against binary and multi-class ANNs.}

\textcolor{black}{The label flipping SVM experiment's threat model was white-box such that an adversary could control the input data used to train the model. The~hypothesis predicted that those labels with the smallest margin would significantly affect how the SVM hyperplane was calculated. The~result showed that a significant amount of labels had to be manipulated to affect the accuracy and, most importantly, the~recall score. On~the other hand, using the random flip method, the~accuracy and recall scores could be impacted with a significantly lower amount of manipulated labels. Upon~comparing the results to the related experiments, it was concluded that the manipulated labels with a high margin would affect the SVM hyperplane generation more significantly. The~FGSM was used to generate adversarial examples to perturb the data for the ANN.} The experiments performed adversarial FGSMs against both binary and five-class multi-classification ANN models. \textcolor{black}{The threat model defined that the testing dataset would be perturbed using the FGSM using the CleverHans package
in targeted and untargeted approaches.} In the binary classification experiments, it was found that as the epsilon value grew to 1, the~accuracy, recall, precision and F1 scores dropped. \textcolor{black}{In turn, the~results showed that perturbed data was able to {trick} the ANN significantly. The~five-class multi-classification experiments found that as the epsilon value grew to 1, the~accuracy, recall, precision, and~F1 scores dropped. The~increasing epsilon value perturbs the data more significantly, which resulted in more significantly impacted metrics.}

\textcolor{black}{Our results show that the Bot-IoT classes are imbalanced, something that is addressed by the literature by configuring the models' class weighting parameter.}
\textcolor{black}{ However, adversarial example generation in a dataset with balanced classes remains difficult to create in a real-world environment and is positioned first in our priority list to test as soon as it becomes available. Additionally, the~multi-classification experiments aggregated all types of attacks existing in the Bot-IoT dataset, rather than investigating the specific attacks individually. We also plan to investigate how different adversarial methods could be used for the Bot-IoT dataset's particular attacks. As~our work did not cover manipulating labels with a high margin to the SVM hyperplane, we also want to explore further how these labels affect the SVM model. Finally, we want to implement and study the suggested countermeasures' effects on the adversarial machine learning methods investigated.}
\vspace{6pt}

\authorcontributions{All authors contributed in the conceptualization and methodology of the manuscript; O.T.v.E. performed the data preparation; P.P., N.P. and C.C. contributed in writing; A.M. and W.J.B. reviewed and edited the manuscript. All authors have read and agreed to the published version of the~manuscript.}

\funding{This research received no external funding.}

\conflictsofinterest{The authors declare no conflict of~interest.}
\newpage

\appendixtitles{yes}
\appendixstart
\appendix
\makeatletter
\setcounter{lstlisting}{0}
\renewcommand{\thelstlisting}{A\arabic{lstlisting}}

\section{SVM Random Label Flip 5\% Sample}
\label{app:SVMrandomflip}
\vspace{-5pt}

\begin{center}
\lstset{%
caption=Descriptive Caption Text,
basicstyle=\ttfamily\footnotesize\bfseries,
frame=tb
}
\begin{lstlisting}[language=python, caption=SVM random label flip 5\% sample., label=code:svm-random-flip]
n = 0.05 # 5%
change = training.sample(int(n*len(training))).index
# Use binary XOR to flip from 1 to 0 and vice versa
training.loc[change, 'attack'] ^= 1
\end{lstlisting}
\end{center}

\section{SVM Targeted Label Flip~Function}
\label{appF}
\label{app:SVMtargetedflip}
\vspace{-5pt}
\begin{center}
\lstset{%
caption=Descriptive Caption Text,
basicstyle=\ttfamily\footnotesize\bfseries,
frame=tb
}
\begin{lstlisting}[language=python, caption=SVM targeted label flip function., label=code:svm-targeted-flip]
def get_change(model, X_train, n):
    distances = model.decision_function(X_train)
    abs_values = np.abs(distances) # gen absolute distance
    df = pd.DataFrame(data={'distance': distances, 
                            'abs_value': abs_values}, 
                      columns=["distance", "abs_value"])
    
    change = df.sort_values(
        by=['abs_value']).head(int(n*len(df))).index
    
    return change
\end{lstlisting}
\end{center}

\section{Create Logits~Model}
\label{app:logits}
\vspace{-5pt}
\begin{center}
\lstset{%
caption=Descriptive Caption Text,
basicstyle=\ttfamily\footnotesize\bfseries,
frame=tb
}
\begin{lstlisting}[language=python, caption=Create logits model., label=code:logits-model]
import tensorflow as tf # Version 2.3.0
logits_model = tf.keras.Model(model.input,
                              model.layers[-1].output)
\end{lstlisting}
\end{center}

\section{Convert Testing Dataset into~Tensor}
\label{app:converttensor}
\vspace{-5pt}
\begin{center}
\lstset{%
caption=Descriptive Caption Text,
basicstyle=\ttfamily\scriptsize\bfseries,
frame=tb
}
\begin{lstlisting}[language=python, caption=Convert testing dataset into tensor., label=code:create-X-tensor]
original_data = X_test_scaled
original_data = tf.convert_to_tensor(original_data.reshape((len(X_test_scaled), 10)))
\end{lstlisting}
\end{center}

\section{Use Fast Gradient Sign Method to Generate Adversarial~Examples}
\label{app:fgsm}
\vspace{-5pt}
\begin{center}
\lstset{%
caption=Descriptive Caption Text,
basicstyle=\ttfamily\footnotesize\bfseries,
frame=tb
}
\begin{lstlisting}[language=python, caption=Use fast gradient sign method to generate adversarial examples., label=code:fgsm-adv-example]
import cleverhans.future.tf2.attacks.fast_gradient_method  

epsilon = 1
adv_data = fast_gradient_method(logits_model, original_data, 
                                epsilon, np.inf, 
                                targeted=False)
                                
adv_data_pred = model.predict(adv_data)
\end{lstlisting}
\end{center}

\section{Visualisation of~Metrics}
\label{app:metricsplots}

\textcolor{black}{The metric results were visualised using Matplotlib Pyplot (Matplotlib: Visualization with Python: \url{https://matplotlib.org/} (accessed on 22 April 2021)).}

\begin{figure}[H]

\begin{tabular}{cc}
\includegraphics[width=0.4\linewidth]{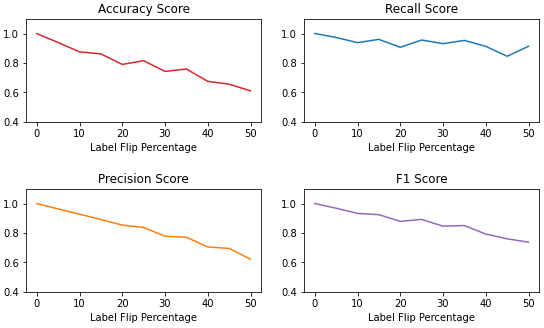}
& \includegraphics[width=0.4\linewidth]{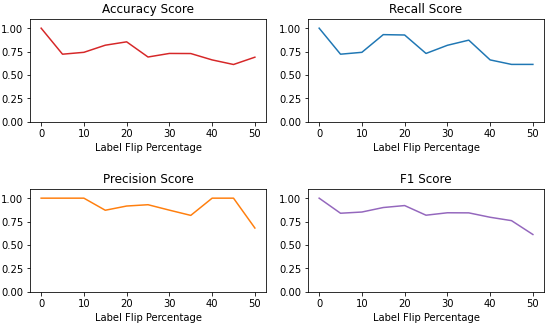}\\
({\bf a})&({\bf b})\\
\end{tabular}
\caption{(\textbf{a}) {Targeted Support Vector Machines flip metrics}. (\textbf{b}) {Non-targeted Support Vector Machines flip metrics.}}
\label{fig:svm_targeted_non_targeted}
\end{figure}
\vspace{-5pt}

\begin{figure}[H]

\begin{tabular}{cc}
\includegraphics[width=0.4\linewidth]{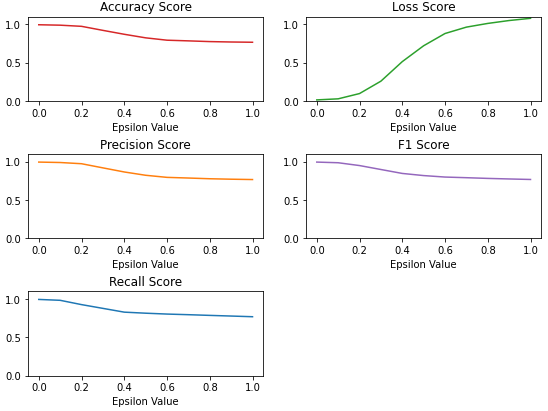}
& \includegraphics[width=0.4\linewidth]{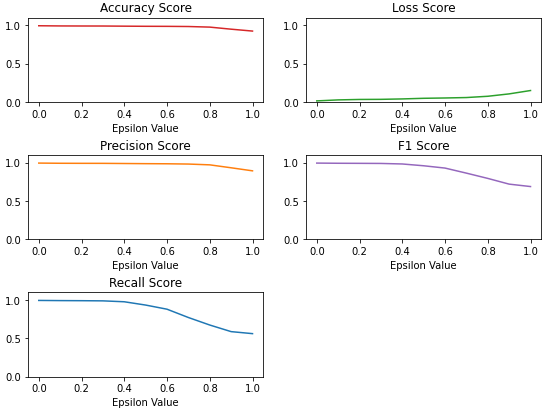}\\
({\bf a})&({\bf b})\\
\end{tabular}

\caption{(\textbf{a}) {Binary classification metrics, non-targeted Fast Gradient Sign Method.} (\textbf{b}) {Binary classification metrics, targeted Fast Gradient Sign Method.}}
\label{fig:binary_non_targeted_targeted}
\end{figure}
\vspace{-5pt}

\begin{figure}[H]

\begin{tabular}{cc}
\includegraphics[width=0.4\linewidth]{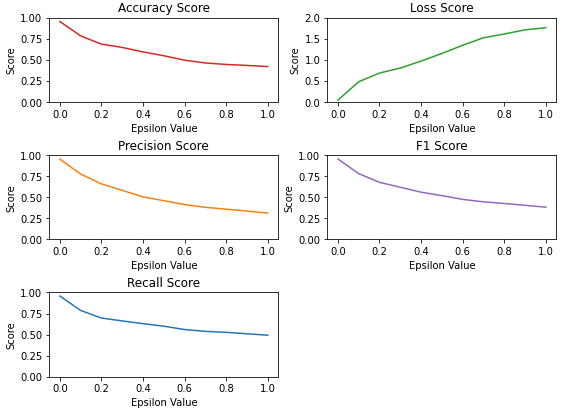}
& \includegraphics[width=0.4\linewidth]{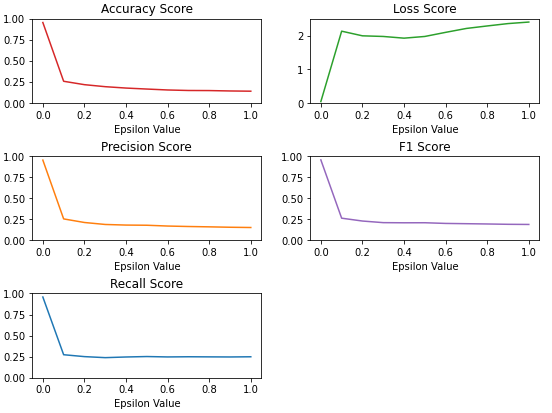}\\
({\bf a})&({\bf b})\\
\end{tabular}

\caption{(\textbf{a}) {Multi-classification metrics, targeted Fast Gradient Sign Method}. (\textbf{b}) {Multi-classification metrics, non-targeted Fast Gradient Sign Method}.}
\label{fig:multi_targeted_non_targeted}
\end{figure}
\unskip

\section{Software and~Hardware}
\label{app:software_details}

\textcolor{black}{Many tools can be used to process the Bot-IoT dataset. Initially, the~Python programming language was selected due to the abundance of packages that have been developed, namely the Scikit-Learn package~\cite{srinath2017python}. \textcolor{black}{The focus of the Scikit-Learn package is to create an approachable interface while providing state-of-the-art machine learning algorithm implementations~\cite{pedregosa2011scikit}.} Conversely, Scikit-Learn has some limitations; for example, it may not be scalable in performance when considering large datasets. To~combat the limitations of Scikit-Learn, TensorFlow has been used to spread the workload across many different cores~\cite{abadi2016tensorflow}. TensorFlow is an open-source project developed by Google that is popular within the data science community for its flexibility in maximizing the use of the available hardware. TensorFlow's primary benefit is the inclusion of Keras, a~library used to develop neural networks in Python, thus allowing data analysis using estimators~\cite{abadi2016tensorflow}.}

\textcolor{black}{As it can be seen in Figure~\ref{testbed-architecture}, the~system's configuration is provisioned through the Google Colaboratory (Colab) (Google Colaboratory: \url{https://colab.research.google.com/} (accessed on 22 April 2021)) service. Google Colab is a cloud Software as a Service (SaaS) platform. The~GPU provided in Google Colab is an Nvidia Tesla K80 which has 4992 CUDA cores. The~high amount of CUDA cores will be beneficial for the machine learning model training using TensorFlow.}
\vspace{-4pt}
\begin{figure}[H]
\includegraphics[width=0.7\linewidth]{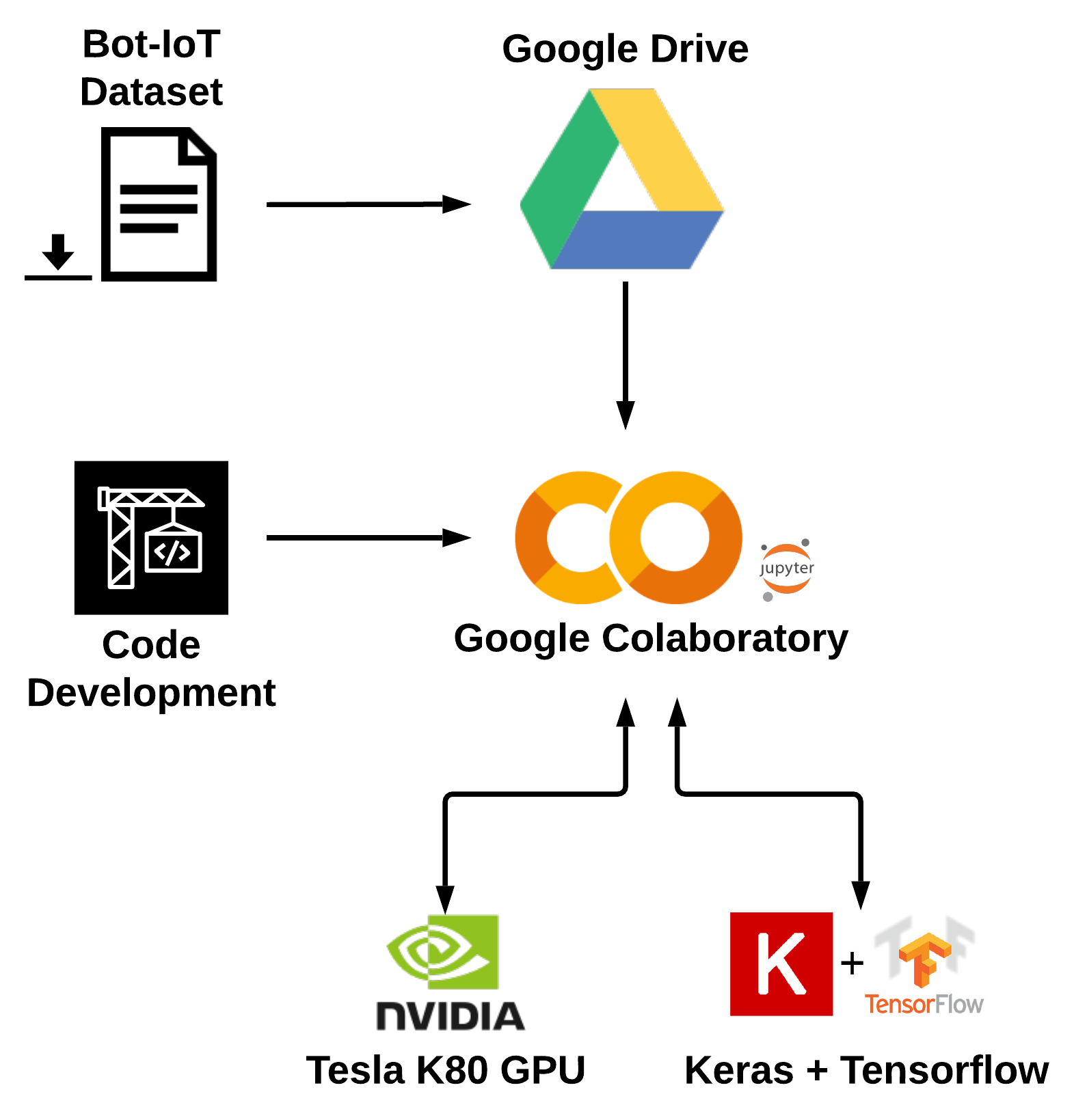}
\caption{\textcolor{black}{Testbed architecture}.}
\label{testbed-architecture}
\end{figure}

\end{paracol}
\reftitle{References}

\end{document}